\documentclass[11pt,a4paper]{article}
\usepackage[nohyperref]{acl2019}
\usepackage{times}
\usepackage{latexsym}
\usepackage{url}
\usepackage{graphicx}
\usepackage{subcaption}
\usepackage{booktabs}
\usepackage{soul}
\usepackage{tcolorbox}
\definecolor{itcolor}{rgb}{0.622, 0.135, 0.198}
\definecolor{stcolor}{rgb}{0.100, 0.835, 0.328}

\newtcbox{\itbox}{on line,
  colframe=itcolor,colback=itcolor!10!white,
  boxrule=0.5pt,arc=4pt,boxsep=0.5pt,left=2pt,right=2pt,top=2pt,bottom=2pt}

\newtcbox{\stbox}{on line,
  colframe=stcolor,colback=stcolor!10!white,
  boxrule=0.5pt,arc=4pt,boxsep=0.5pt,left=2pt,right=2pt,top=2pt,bottom=2pt}

\usepackage{floatrow}
\newfloatcommand{capbtabbox}{table}[][]

\usepackage{array}
\newcommand{\PreserveBackslash}[1]{\let\temp=\\#1\let\\=\temp}
\newcolumntype{C}[1]{>{\PreserveBackslash\centering}p{#1}}
\newcolumntype{R}[1]{>{\PreserveBackslash\raggedleft}p{#1}}
\newcolumntype{L}[1]{>{\PreserveBackslash\raggedright}p{#1}}

\newcommand{\vecx}{{\ensuremath{\bf{x}}}}
\newcommand{\T}{\ensuremath{\mathcal{T}}}

\aclfinalcopy

\title{Undersensitivity in Neural Reading Comprehension}
\author{
  Johannes Welbl{$^\dagger$} \quad Pasquale Minervini{$^\dagger$} \quad Max Bartolo{$^\dagger$} \\
  \textbf{Pontus Stenetorp}{$^\dagger$} \quad \textbf{Sebastian Riedel}{$^{\ddagger\dagger}$}\\
  {$^\dagger$}University College London \quad {$^{\ddagger}$}Facebook AI Research\\
  { \normalsize \tt \{j.welbl,p.minervini,m.bartolo,p.stenetorp,s.riedel\}@cs.ucl.ac.uk}
}

\date{}
\begin{document}
\maketitle

\begin{abstract}
Current reading comprehension models generalise well to in-distribution test sets, yet perform poorly on adversarially selected inputs.
Most prior work on adversarial inputs studies oversensitivity: semantically invariant text perturbations that cause a model's prediction to change when it should not.
In this work we focus on the complementary problem: excessive prediction undersensitivity, where input text is meaningfully changed but the model's prediction does not, even though it should. 
We formulate a noisy adversarial attack which searches among semantic variations of the question for which a model erroneously predicts the same answer, and with even higher probability.
Despite comprising unanswerable questions, both \emph{SQuAD2.0} and \emph{NewsQA} models are vulnerable to this attack.
This indicates that although accurate, models tend to rely on spurious patterns and do not fully consider the information specified in a question.
We experiment with data augmentation and adversarial training as defences, and find that both substantially decrease vulnerability to attacks on held out data, as well as held out attack spaces.
Addressing undersensitivity also improves results on \textsc{AddSent} and \textsc{AddOneSent}, and models furthermore generalise better when facing train/evaluation distribution mismatch: 
they are less prone to overly rely on predictive cues present only in the training set, and outperform a conventional model by as much as 10.9\%~F$_1$.
\end{abstract}

\begin{figure*}[t]
    \centering
    \includegraphics[width=0.85\linewidth]{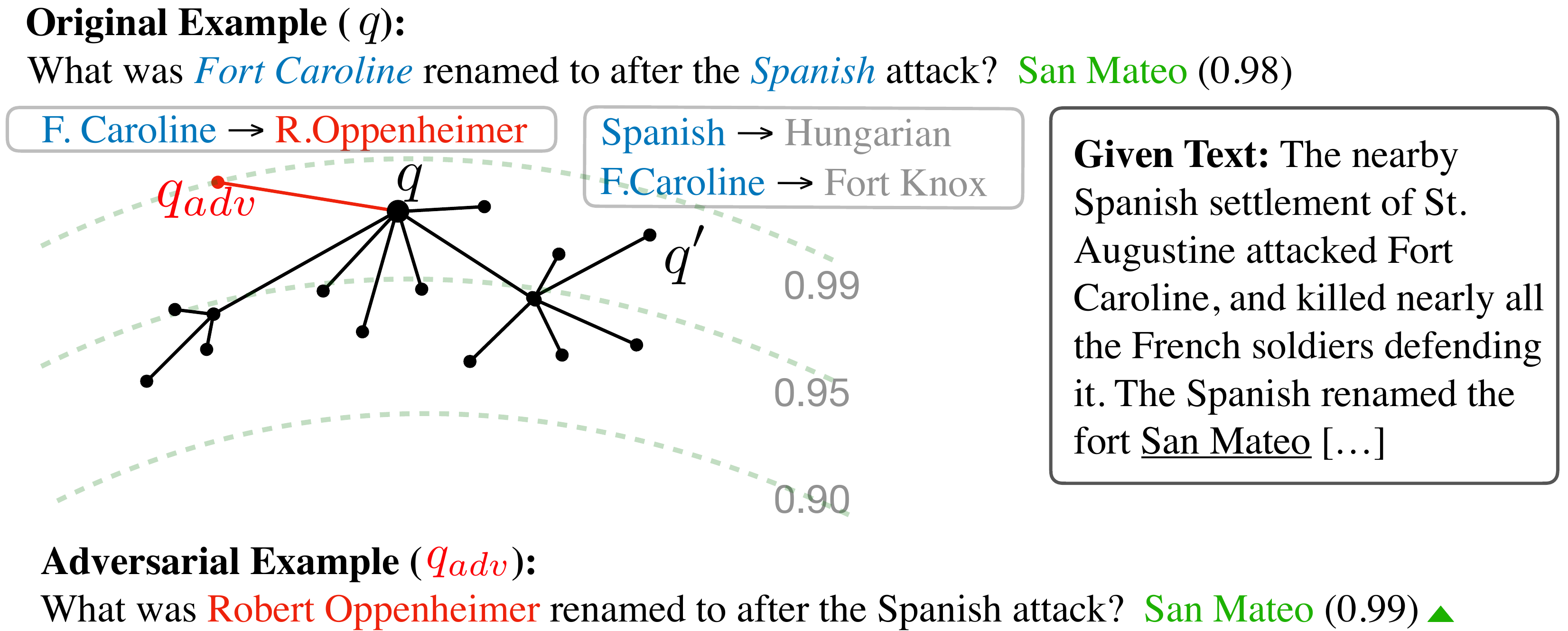}
    \caption{Method Overview: Adversarial search over semantic variations of RC questions, producing unanswerable questions for which the model retains its predictions with even higher probability.} 
    \label{fig:overview}
\end{figure*}

\section{Introduction}
Neural networks can be vulnerable to adversarial input perturbations~\citep{szegedy2013intriguing,kurakin2016adversarial}. 
In Natural Language Processing (NLP), which operates on discrete symbol sequences, adversarial attacks can take a variety of forms~\citep{Ettinger17,alzantot2018generating} including character perturbations~\citep{ebrahimi2018hotflip}, semantically invariant reformulations~\citep{ribeiro2018semantically,iyyer2018syntactically} 
or -- specifically in Reading Comprehension (RC) -- adversarial text insertions~\citep{jia2017adversarial,wang2018robust}. 
A model's inability to handle adversarially chosen input text puts into perspective 
otherwise impressive generalisation results for in-distribution test sets (\citet{seo2017bidaf,yu2018qanet,devlin2019bert}; \emph{inter alia}) and constitutes an important caveat to conclusions drawn regarding a model's language understanding abilities.

While semantically invariant text transformations can remarkably alter a model's predictions, the converse problem of model \emph{undersensitivity} is equally troublesome: a model's text input can often be drastically changed in meaning while retaining the original prediction. In particular, previous works~\citep{feng2018pathologies,Ribeiro2018anchors,welbl2020towards} show that even after deletion of all but a small fraction of input words, models often produce the same output. However, such reduced inputs are usually unnatural to a human reader, and it is both unclear what behaviour we should expect from natural language models evaluated on unnatural text, and how to use such unnatural inputs to improve models. In this work we show that, in RC, undersensitivity can be probed with automatically generated natural language questions. In turn, we use these to both make RC models more sensitive when they should be, and more robust in the presence of biased training data.

Figure~\ref{fig:overview} shows an example for a \textsc{BERT Large} model~\citep{devlin2019bert} trained on \emph{SQuAD2.0}~\citep{rajpurkar2018squad2} that is given a text and a comprehension question, i.e.~\emph{What was Fort Caroline renamed to after the Spanish attack?} which it correctly answers as \emph{San Mateo} with 98\% confidence.
Altering this question, however, can increase model confidence for this same prediction to 99\%, even though the new question is unanswerable given the same context. That is, we observe an increase in model confidence, despite removing relevant question information and replacing it with irrelevant content.

We formalise the process of finding such questions as an adversarial search in a discrete input space arising from perturbations of the original question.
There are two types of discrete perturbations we consider, based on part-of-speech and named entities, with the aim of obtaining grammatical and semantically consistent alternative questions that do not accidentally have the same correct answer.
We find that \emph{SQuAD2.0} and \emph{NewsQA}~\cite{trischler2017newsqa} models can be attacked on a substantial proportion of samples, even with a limited computational adversarial search budget.

The observed undersensitivity correlates negatively with in-distribution test set performance metrics (EM/F$_1$), suggesting that this phenomenon -- where present -- is a reflection of a model's lack of question comprehension.
When training models to defend against undersensitivity attacks with data augmentation and adversarial training, we observe that they can generalise their robustness to held out evaluation data without sacrificing in-distribution test set performance.
Furthermore, the models improve on the adversarial datasets proposed by \newcite{jia2017adversarial}, and behave more robustly in a learning scenario that has dataset bias with a train/evaluation distribution mismatch, increasing performance by up to 10.9\%F$_1$.
In summary, our contributions are as follows:
\begin{itemize}
    \item We propose a new type of adversarial attack targeting the undersensitivity of neural RC models, and show that current models are vulnerable to it.
    \item We compare two defence strategies, data augmentation and adversarial training, and show their effectiveness at reducing undersensitivity errors on held-out data and held-out perturbations, without sacrificing in-distribution test set performance.
    \item We demonstrate that the resulting models generalise better on \mbox{\textsc{AddSent}} and \mbox{\textsc{AddOneSent}} \cite{jia2017adversarial}, as well as in a biased data scenario, improving their ability to answer questions with many possible answers when trained on questions with only one.
\end{itemize}

\section{Related Work}
\paragraph{Adversarial Attacks in NLP}
Adversarial examples have been studied extensively in NLP -- see \citet{zhang2019survey} for a recent survey.
However, automatically generating adversarial examples in NLP is non-trivial, as the search space is discrete and altering a single word can easily change the semantics of an instance or render it incoherent.
Recent work overcomes this issue by focusing on simple semantic-invariant transformations, showing that neural models can be \emph{oversensitive} to such modifications of the inputs.
For instance, \citet{ribeiro2018semantically} use a set of simple perturbations such as replacing \emph{Who is} with \emph{Who's}.
Other semantics-preserving perturbations include typos~\citep{DBLP:conf/cvpr/HosseiniXP17}, the addition of distracting sentences~\citep{jia2017adversarial,wang2018robust}, character-level adversarial perturbations~\citep{ebrahimi2018hotflip}, and paraphrasing~\citep{DBLP:conf/naacl/IyyerWGZ18}.

In this work, we instead focus on \emph{undersensitivity} of neural RC models to semantic perturbations of the input.
This is related to previous works leveraging domain knowledge for the generation of adversarial examples~\citep{DBLP:conf/acl/HovyKSK18,DBLP:conf/conll/Minervini018}: our method is based on the idea that modifying, for instance, the named entities involved in a question can completely change its meaning and, as a consequence, the answer to the question should also differ.
Our approach does not assume white-box access to the model, as do e.g.~\citet{ebrahimi2018hotflip} and \citet{wallace2019universal}.

\paragraph{Undersensitivity}
\citet{jacobsen2018excessive} demonstrated classifier undersensitivity in computer vision, where altered input images can still produce the same prediction scores, achieved using (approximately) invertible networks.
\citet{niu2018adversarial} investigated over- and undersensitivity in dialogue models and addressed the problem with a max-margin training approach.
\citet{Ribeiro2018anchors} describe a general model diagnosis tool to identify minimal feature sets that are sufficient for a model to form high-confidence predictions.
\citet{feng2018pathologies} showed that it is possible to reduce inputs to minimal input word sequences without changing a model's predictions.
\citet{welbl2020towards} investigated formal verification against undersensitivity to text deletions.

We see our work as a continuation of this line of inquiry, but with a particular focus on undersensitivity in RC.
In contrast to prior work \cite{feng2018pathologies,welbl2020towards}, we consider concrete alternative questions, rather than arbitrarily reduced input word sequences.
We furthermore address the observed undersensitivity using dedicated training objectives, in contrast to \citet{feng2018pathologies} and \citet{Ribeiro2018anchors} who simply highlight it.

Finally, one of the baseline methods we later test for defending against undersensitivity attacks is a form of data augmentation that has similarly been used for de-biasing NLP models \citep{zhao2018gender,Lu2018Gender}.

\paragraph{Unanswerable Questions in Reading Comprehension}
Following \citet{jia2017adversarial}'s publication of adversarial attacks on the \emph{SQuAD1.1} dataset, \citet{rajpurkar2018squad2} proposed the \emph{SQuAD2.0} dataset, which includes over 43,000 human-curated unanswerable questions.
A second dataset with unanswerable question is \emph{NewsQA}~\citep{trischler2017newsqa}, comprising questions about news texts.
Training on these datasets should conceivably result in models with an ability to tell whether questions are answerable or not; we will see, however, that this does not extend to adversarially chosen unanswerable questions in our undersensitivity attacks.
\citet{Hu2019ReadVerify} address unanswerability of questions from a given text using additional verification steps. Other approaches have shown the benefit of synthetic data to improve performance in \emph{SQuAD2.0} \citep{zhu2019learning,DBLP:conf/acl/AlbertiAPDC19}.

We operate on the same underlying research premise, that the ability to handle unanswerable questions is an important part of improving text comprehension models.
In contrast to prior work, we demonstrate that despite improving performance on test sets that include unanswerable questions, the problem persists when adversarially choosing from a larger space of questions.

\section{Methodology}
\paragraph{Problem Overview}
Consider a discriminative model $f_{\theta}$, parameterised by a collection of vectors $\theta$, which transforms an input $\vecx$ into a prediction $\hat y=f_{\theta}(\vecx)$.
In our task, $\vecx=(t,q)$ is a given text $t$ paired with a question $q$ about this text. The label $y$ is the answer to $q$ where it exists, or a \emph{NoAnswer} label where it cannot be answered.\footnote{Unanswerable questions are part of, e.g.~the \emph{SQuAD2.0} and \emph{NewsQA} datasets, but not of \emph{SQuAD1.1.}}

In a text comprehension setting the set of possible answers is large, and predictions $\hat{y}$ should be dependent on $\vecx$. 
And indeed, randomly choosing a different input $(t', q')$ is usually associated with a change of the model prediction $\hat{y}$.
However, there exist many examples where the prediction erroneously remains stable;
the goal of the attack formulated here is to find such cases.
Concretely, given a computational search budget, the goal is to discover inputs $\vecx'$, for which the model still erroneously predicts $f_{\theta}(\vecx') = f_{\theta}(\vecx)$, even though $\vecx'$ is not answerable from the text.

\paragraph{Input Perturbation Spaces}
Identifying suitable candidates for $\vecx'$ can be achieved in manifold ways. 
One approach is to search among a large question collection, but we find this to only rarely be successful; an example is shown in Table~\ref{tab:alternative_question}, Appendix~\ref{apdx:example}.
Generating $\vecx'$, on the other hand, is prone to result in ungrammatical or otherwise ill-formed text. 
Instead, we consider a perturbation space $\mathcal{X}_\T(\vecx)$ spanned by perturbing original inputs $\vecx$ using a perturbation function family~$\T$:
\begin{equation}
    \mathcal{X}_\T(\vecx) = \{T_i(\vecx) \mid T_i \in \T\}
\end{equation}
This space $\mathcal{X}_\T(\vecx)$ contains alternative model inputs derived from \vecx. Ideally the transformation function family \T~is chosen such that the correct label of these new inputs is changed: for $\vecx' \in \mathcal{X}_\T(\vecx): y(\vecx') \neq y(\vecx)$.
We will later search within $\mathcal{X}_\T(\vecx)$ to find inputs $\vecx'$ which erroneously retain the same prediction as \vecx: $\hat y(\vecx)=\hat y(\vecx')$.
%
\paragraph{Part-of-Speech (PoS) Perturbations}
We first consider the perturbation space $\mathcal{X}_{\T_P}(\vecx)$ generated by PoS perturbations $\T_P$ of the original question:
we swap individual tokens with other, PoS-consistent alternative tokens, where we draw from large collections of tokens of the same PoS types. 
For example, we might alter the question \emph{Who patronized the monks in Italy?} to \emph{Who betrayed the monks in Italy?} by replacing the past tense verb \emph{patronized} with \emph{betrayed}.
There is however no guarantee that the altered question will require a different answer (e.g.~due to synonyms). 
Even more so -- there might be type clashes or other semantic inconsistencies (e.g.~\emph{Who built the monks in Italy?}).
We perform a qualitative analysis to investigate the extent of this problem and find that, while a valid concern, for the majority of attackable samples there exist attacks based on correct well-formed questions (see Section \ref{sec:analysis}).
%
\paragraph{Named Entity Perturbations}
The space $\mathcal{X}_{\T_E}(\vecx)$ generated by the transformation family $\T_E$ is created by substituting mentions of named entities (NE) in the question with different type-consistent NE, derived from a large collection $E$.
For example, a comprehension question \emph{Who patronized the monks in Italy?} could be altered to \emph{Who patronized the monks in Las Vegas?}, replacing the geopolitical entity \emph{Italy} with \emph{Las Vegas}, chosen from $E$.
Altering NE often changes the specifics of the question and poses different requirements to the answer, which are unlikely to be satisfied from what is stated in the given text, given the broad nature of entities in $E$.
While perturbed questions are not guaranteed to be unanswerable or require a different answer, we will later find in a qualitative analysis that for the large majority of cases they do.

\begin{figure*}[t]
    \begin{subfigure}{0.42\textwidth}
    \includegraphics[width=\linewidth]{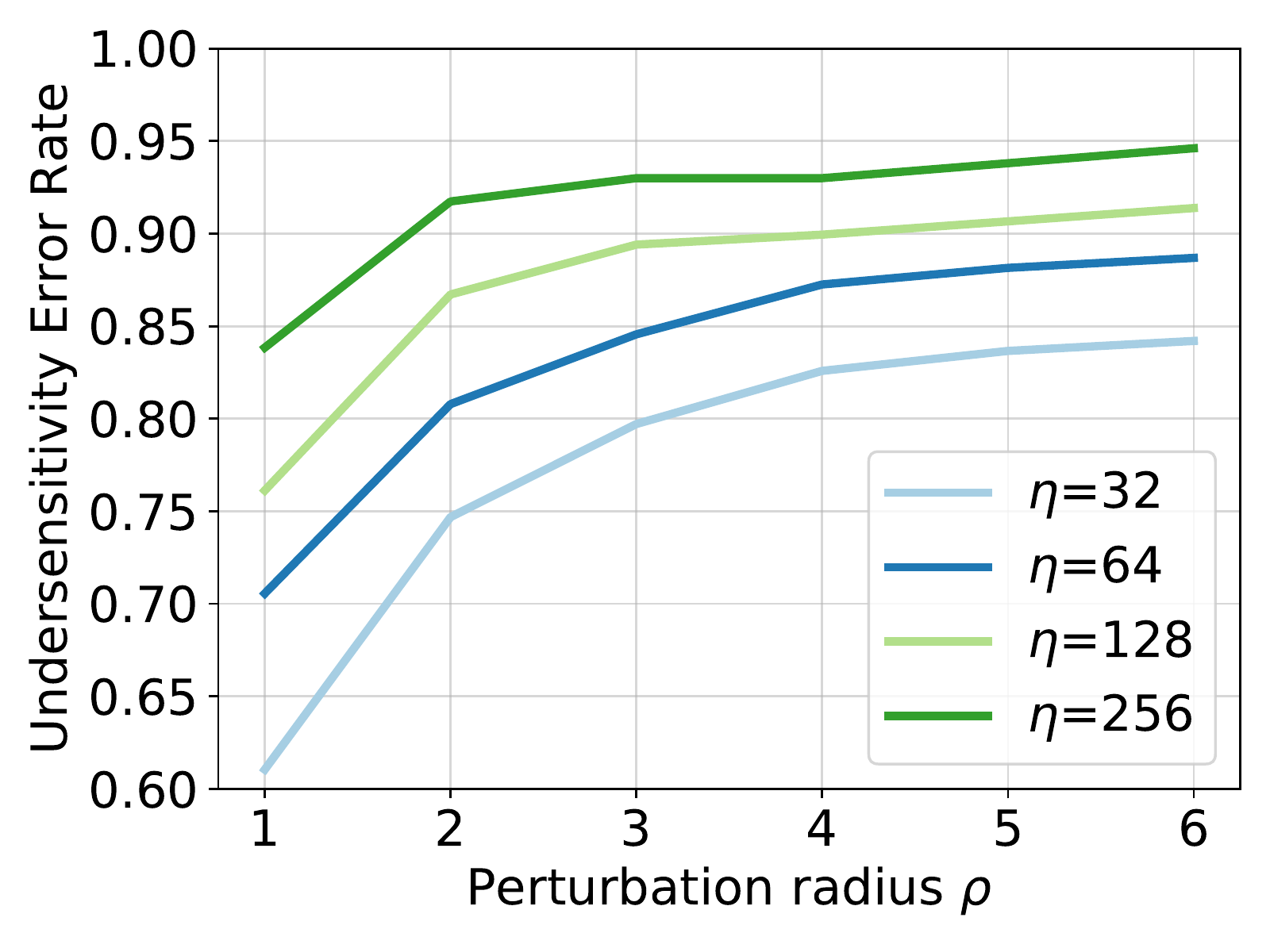}
    \caption{
        Part of Speech perturbations
    }  \label{subfig:a}
    \end{subfigure}
    \hspace{0.1\textwidth}
    \begin{subfigure}{0.42\textwidth}
    \includegraphics[width=\linewidth]{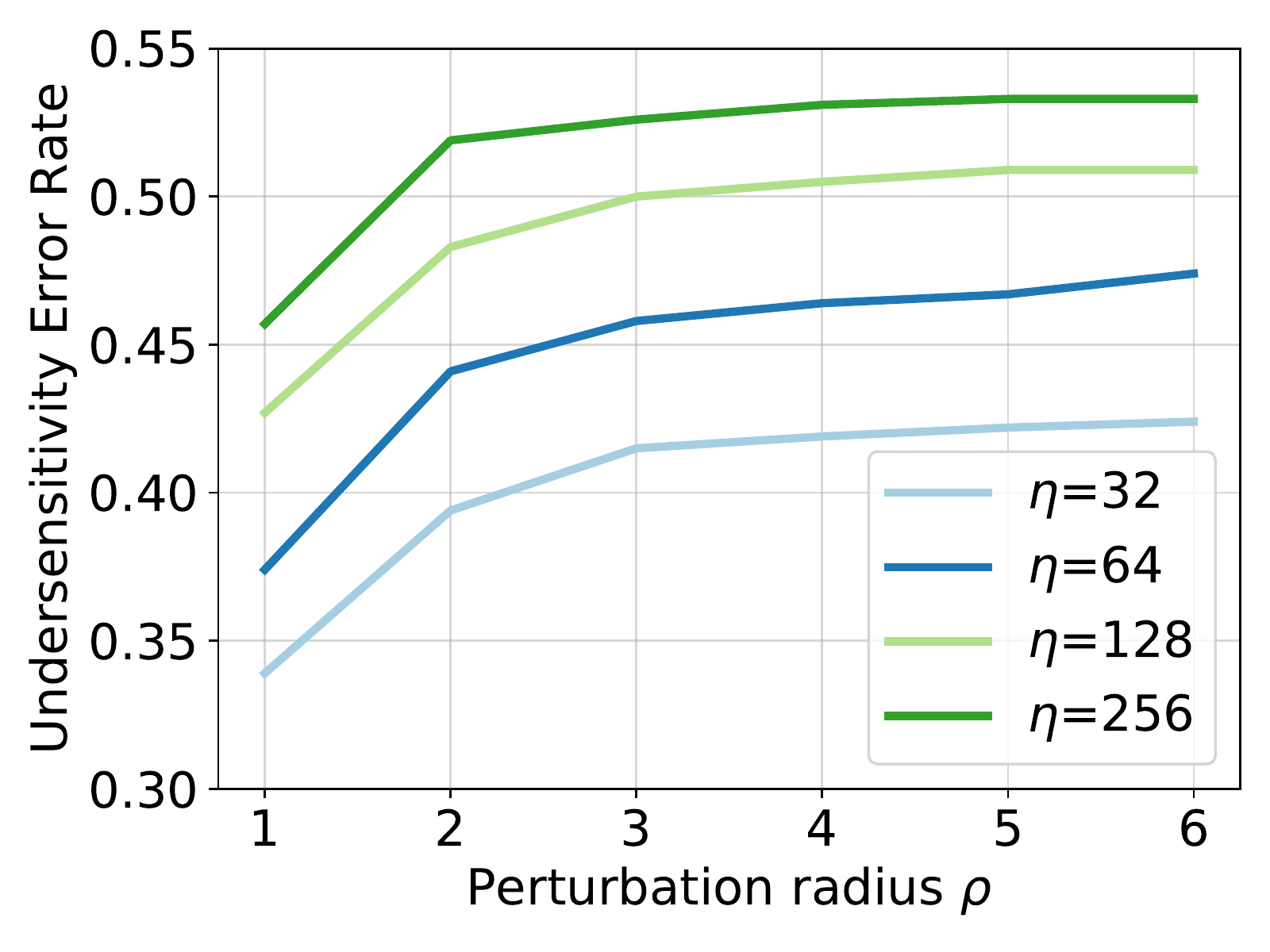}
    \caption{
        Named Entity perturbations 
    }  \label{subfig:1b}
    \end{subfigure}
    \caption{\textsc{BERT Large} on \emph{SQuAD2.0}: vulnerability to noisy attacks on held out data for differently sized attack spaces (parameter $\eta$) and different beam search depth (perturbation radius $\rho$).} \label{fig:2}
    \label{fig:squad2_vulnerability}
\end{figure*}

\paragraph{Undersensitivity Attacks}
Thus far we have described different methods of perturbing questions.
We will search in the resulting perturbation spaces $\mathcal{X}_{\T_P}(\vecx)$ and $\mathcal{X}_{\T_E}(\vecx)$ for inputs $\vecx'$ for which the model prediction remains constant.
However, we pose a slightly stronger requirement: $f_{\theta}$ should assign an even higher probability to the same prediction $\hat{y}(\vecx)=\hat{y}(\vecx')$ than for the original input:
\begin{equation}\label{eq:undersensitivity}
P(\hat{y} \mid \vecx ') > P(\hat{y} \mid \vecx)
\end{equation}
This defines a sufficient criterion for preserving the prediction.
It is a conservative choice, as it includes only the subset of prediction-preserving cases with \emph{increased} model probability.

To summarise, we are searching in a perturbation space for altered questions which result in a higher model probability to the same answer as the original input question.
If we have found an altered question that satisfies inequality~(\ref{eq:undersensitivity}), then we have identified a successful attack, which we will refer to as an \emph{undersensitivity attack}.

\paragraph{Adversarial Search in Perturbation Space}
In its simplest form, a search for an adversarial attack in the previously defined attack spaces amounts to a search over a list of single lexical alterations for the maximum (or any) higher prediction probability.
We can however repeat the replacement procedure multiple times, arriving at texts with potentially larger lexical distance to the original question.
For example, in two iterations of PoS-consistent lexical replacement, we can alter \emph{Who was the duke in the battle of Hastings?} to inputs like \emph{Who was the duke in the expedition of Roger?}

The space of possibilities with increasing distance grows combinatorially, and with increasing perturbation radius it becomes computationally infeasible to comprehensively cover the full perturbation spaces arising from iterated substitutions.
To address this, we follow \newcite{feng2018pathologies} and apply beam search to narrow the search space, and seek to maximise the difference
\begin{equation}
\Delta = P(\hat{y} \mid \vecx ') - P(\hat{y} \mid \vecx)
\end{equation}
Beam search is conducted up to a pre-specified maximum perturbation radius $\rho$, but once $\vecx'$ with $\Delta>0$ has been found, we stop the search.

\paragraph{Relation to Attacks in Prior Work}
Note that this type of attack stands in contrast to other attacks based on small, \emph{semantically invariant} input perturbations \citep{Belinkov_Bisk2017,ebrahimi2018hotflip,ribeiro2018semantically} which investigate oversensitivity problems.
Such semantic \emph{invariance} comes with stronger requirements and relies on synonym dictionaries \citep{ebrahimi2018hotflip} or paraphrases harvested from back-translation \citep{iyyer2018syntactically}, which are both incomplete and noisy.
Our attack is instead focused on \emph{undersensitivity}, i.e.~where the model is stable in its prediction even though it should not be.
Consequently the requirements are not as difficult to fulfil when defining perturbation spaces that \emph{alter} the question meaning, and one can rely on sets of entities and PoS examples automatically extracted from a large text collection.

In contrast to prior attacks~\citep{ebrahimi2018hotflip,wallace2019universal}, we evaluate each perturbed input with a standard forward pass rather than using a first-order Taylor approximation to estimate the output change induced by a change in input. 
This is less efficient but exact, and furthermore does not require white-box access to the model and its parameters.

\section{Experiments: Model Vulnerability}
\paragraph{Training and Dataset Details}
We next conduct experiments using the attacks laid out above to investigate model undersensitivity.
We attack the \textsc{BERT} model \citep{devlin2019bert} fine-tuned on \emph{SQuAD2.0}~\citep{rajpurkar2018squad2}, and measure to what extent the model exhibits undersensitivity when adversarially choosing input perturbations.
Our choice of \textsc{BERT} is motivated by its currently widespread adoption across the NLP field, and empirical success across a wide range of datasets.

Note that \emph{SQuAD2.0} per design contains unanswerable questions in both training and evaluation sets; models are thus trained to predict a \emph{NoAnswer} option where a comprehension question cannot be answered.

In a preliminary pilot experiment, we first train a \textsc{BERT Large} model on the full training set for 2 epochs, where it reaches 78.32\%EM and 81.44\%F$_1$, in close range to results (78.7\%EM and 81.9\%F$_1$) reported by \citet{devlin2019bert}.
We then however choose a different training setup as we would like to conduct adversarial attacks on data inaccessible during training: we split off 5\% from the original training set for development purposes and retain the remaining 95\% for training, stratified by articles.
We use this development data to tune hyperparameters and perform early stopping, evaluated every 5,000 steps with batch size 16 and patience 5, and will later tune hyperparameters for defence on it.
The original \emph{SQuAD2.0} development set is then used as evaluation data, where the model reaches 73.0\%EM and 76.5\%F$_1$; we will compute the undersensitivity attacks on this entirely held out part of the dataset.

\begin{table*}[t]
\begin{center}
\resizebox{\columnwidth}{!}{%
\begin{tabular}{l l l r}
\toprule
{\bf Original / Modified Question}  & \bf{Prediction} & {\bf{Annotation}} & \bf{Scores}\\
\midrule
What city in Victoria is called the cricket ground of  & Melbourne & valid  &   \stbox{0.63}/\itbox{0.75}    \\ 
{\stbox{Australia}} {\itbox{the Delhi Metro Rail Corporation Limited}}? & &  &    \\
\midrule
What are some of the accepted general principles of  & fundamental & valid & \stbox{0.59}/\itbox{0.61}  \\ 
{\stbox{European Union}} {\itbox{Al-Andalus}} law?  &       rights [...] & &   \\
\midrule
What were the {\stbox{annual}} {\itbox{every year}} carriage fees for  & \pounds 30m  & same answer  & \stbox{0.95}/\itbox{0.97}    \\ 
the channels?  && & \\
\midrule
\midrule
What percentage of Victorians are {\stbox{Christian}} {\itbox{Girlish}}? & 61.1\% & valid   & \stbox{0.92}/\itbox{0.93}  \\ 
\midrule
Which plateau is the left {\stbox{part}} {\itbox{achievement}} of Warsaw on?  & moraine  & semantic & \stbox{0.52}/\itbox{0.58}  \\
                                                    &     &inconsistency &\\  
\midrule
 Who leads the {\stbox{Student}} {\itbox{commissioning}} Government?  & an Executive & same answer   & \stbox{0.61}/\itbox{0.65}  \\
 & Committee &  & \\
\bottomrule
\end{tabular}}
\end{center}
\caption{Example adversarial questions (\stbox{original}, \itbox{attack}), together with their annotation as either a valid counterexample or other type.  Top 3: Named Entity perturbations. Bottom 3: PoS perturbations.}\label{tab:labeled_attacks}
\end{table*}

\paragraph{Attack Details}
To compute the perturbation spaces, we collect large sets of string expressions across Named Entity (NE) and PoS types to define the perturbation spaces $\T_E$ and $\T_P$, which we gather from the Wikipedia paragraphs used in the \emph{SQuAD2.0} training set, with the pretrained taggers in \emph{spaCy},\footnote{\url{https://spacy.io}}
and the Penn Treebank tag set for PoS.
This results on average in 5,126 different entities per entity type, and 2,337 different tokens per PoS tag.
When computing PoS perturbations, we found it useful to disregard perturbations of particular PoS types that often led to only minor changes or incorrectly formed expressions, such as punctuation or determiners; more details on the left out tags can be found in Appendix~\ref{apdx:pos}.
As the number of possible perturbations to consider is potentially very large, we limit beam search at each step to a maximum of $\eta$ randomly chosen type-consistent entities from $E$, or tokens from $P$, and re-sample these throughout the search.
We use a beam width of $b=5$, resulting in a bound to the total computation spent on adversarial search of $b \cdot \rho \cdot \eta$  model evaluations per sample, where $\rho$ is the perturbation ``radius'' (the maximum search depth).

\paragraph{Metric: Adversarial Error Rate}
We quantify adversarial vulnerability to the described attacks by measuring the proportion of evaluation samples for which at least one undersensitivity attack is found given a computational search budget, disregarding cases where a model predicts \emph{NoAnswer}.\footnote{Altering unanswerable samples likely retains their unanswerability.}

\subsection{Results}
Figure~\ref{fig:squad2_vulnerability} shows plots for adversarial error rates on \emph{SQuAD2.0} for both perturbation types across various search budgets.
We observe that attacks based on PoS perturbations can already for very small search budgets ($\eta=32$, $\rho=1$) reach more than 60\% attack success rates, and this number can be raised to 95\% with a larger computational budget.
For perturbations based on Named Entity substitution, we find overall lower attack success rates, but still find that more than half of the samples can successfully be attacked under the budgets tested.
Note that where attacks were found, we observed that there often exist multiple alternatives with higher probability. 

These findings demonstrate that \textsc{BERT} is not necessarily considering the entire contents of a comprehension question given to it, and that even though trained to tell when questions are unanswerable, the model often fails when facing adversarially selected unanswerable questions.

In a side experiment we also investigated undersensitivity attacks using Named Entity perturbations on \emph{SQuAD1.1}, which proves even more vulnerable with an adversarial error rate of 70\% already using a budget of $\eta$ = 32; $\rho$ = 1 (compared to 34\% on \emph{SQuAD2.0}). While this demonstrates that undersensitivity is also an issue for \emph{SQuAD1.1}, the unanswerable question behaviour is not really well-defined, rendering results hard to interpret. On the other hand, the notable drop between the datasets demonstrates the effectiveness of the unanswerable questions added during training in \emph{SQuAD2.0}.

\begin{table}[t]
    \noindent
    \centering
    \begin{tabular}{r r r}
        \toprule
                                            &   {\bf PoS}   &   {\bf NE} \\
        \midrule
        Valid attack                        &   51\%        &   84\%      \\
        Syntax error                        &   10\%        &   6\%       \\
        Semantically incoherent             &   24\%        &   5\%       \\
        Same answer                         &   15\%        &   5\%       \\
        \bottomrule
        \end{tabular}
        \captionof{table}{Analysis of undersensitivity attack samples for both PoS and named entity (NE) perturbations.}\label{tab:qualitative}
\end{table}

\begin{figure}[t]
    \centering
    \includegraphics[width=\textwidth]{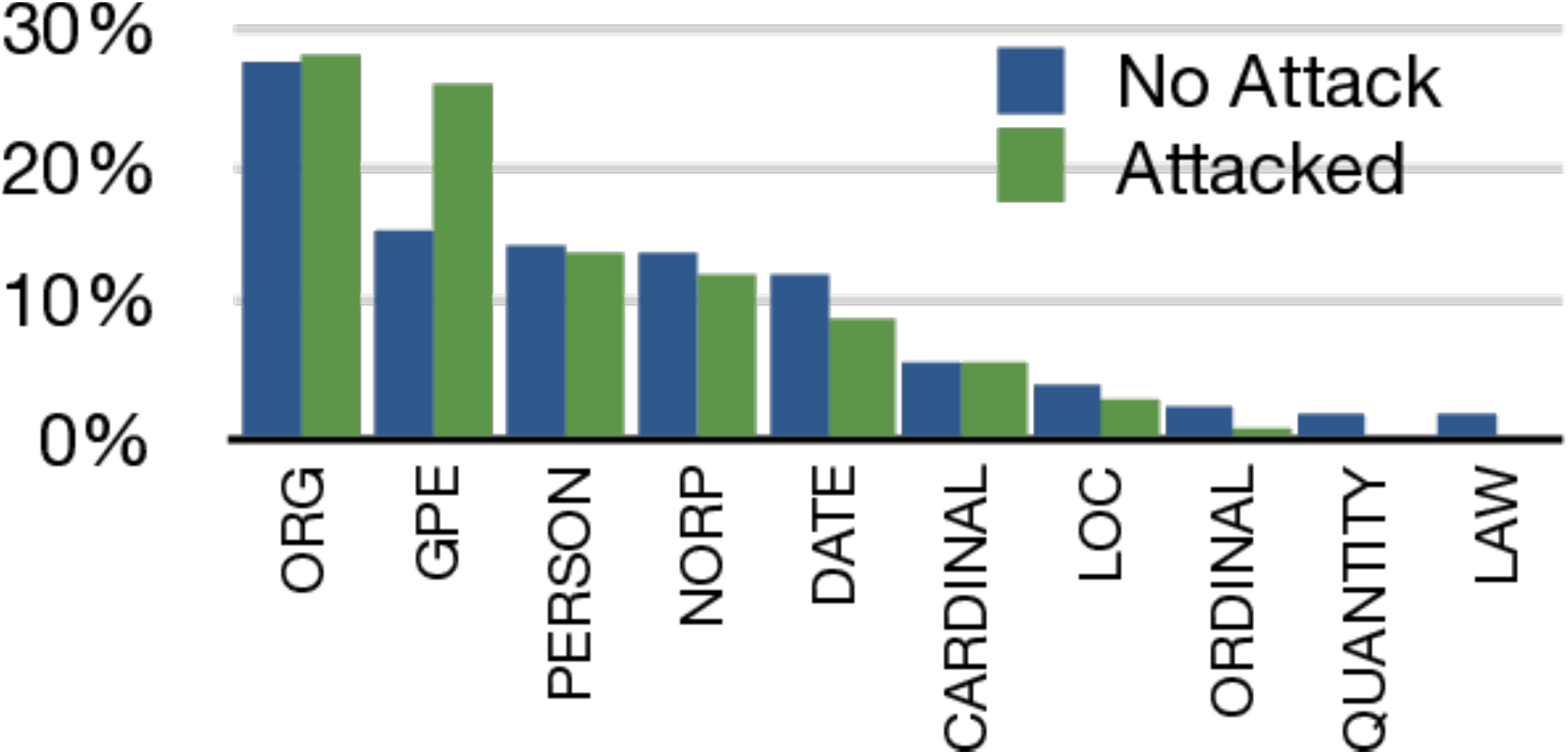}
    \captionof{figure}{Named entity type characteristics of attackable vs.~unattackable samples.} \label{fig:ner_histogram}
\end{figure}

\section{Analysis and Characteristics of Vulnerable Samples}\label{sec:analysis}
\paragraph{Qualitative Analysis of Attacks}
As pointed out before, the attacks are potentially noisy as the introduced substitutions are by no means guaranteed to result in meaningful and semantically consistent expressions, or require a different answer than the original.
To gauge the extent of this we inspect 100 successful attacks conducted at $\rho=6$ and $\eta=256$ on \emph{SQuAD2.0}, both for PoS perturbations and NE perturbations. We label them as either:
\begin{enumerate}
\item Having a syntax error (e.g.~\emph{What would platform lower if there were fewer people?}). These are mostly due to cascading errors stemming from incorrect NE/PoS tag predictions.
\item Semantically incoherent (e.g.~\emph{Who built the monks?}).
\item Questions that require the same correct answer as the original, e.g.~due to a paraphrase. 
\item Valid attacks: Questions that would either demand a different answer or are unanswerable given the text (e.g.~\emph{When did the United States withdraw from the Bretton Woods Accord?} and its perturbed version \emph{When did Tuvalu withdraw from the Bretton Woods Accord?}).
\end{enumerate}

Table~\ref{tab:labeled_attacks} shows several example attacks along with their annotations, and in Table~\ref{tab:qualitative} the respective proportions are summarised.
We observe that a non-negligible portion of questions has some form of syntax error or incoherent semantics, especially for PoS perturbations.
Questions with the identical correct answer are comparatively rare.
Finally, about half (51\%) of all attacks in PoS, as well as 84\% for named entities are valid questions that should either have a different answer, or the \emph{Unanswerable} label.

Overall the named entity perturbations result in much cleaner questions than PoS perturbations, which suffer from semantic inconsistencies in about a quarter of the cases.
While these questions have some sort of inconsistency (e.g.~\emph{What year did the case go before the supreme court?} vs.~a perturbed version \emph{What scorer did the case go before the supreme court?}), it is remarkable that the model assigns higher probabilities to the original answer even when faced with incoherent questions, casting doubt on the extent to which the replaced question information is used to determine the answer.

Since the NE-based attacks have a substantially larger fraction of valid, well-posed alternative questions, we will focus our study on these attacks for the remainder of this paper.

\subsection{Characterising Successfully Attacked Samples}
We have observed that models are vulnerable to undersensitivity adversaries, however not all samples are successfully attacked.
This raises questions regarding what distinguishes samples that can and cannot be attacked.
We investigate various characteristics, aiming to understand model vulnerability causes.

Questions that can be attacked produce lower original prediction probabilities, with an average of $72.9\%$ compared to $83.8\%$ for unattackable questions. 
That is, there exists a direct inverse link between a model's original prediction probability and sample vulnerability to an undersensitivity attack.
The adversarially chosen questions had an average probability of $78.2\%$, i.e.~a notable gap to the original questions. 
It is worth noting that search halted once a single question with higher probability was found; continuing the search increases the respective probabilities.

Vulnerable samples are furthermore less likely to be given the correct prediction overall.
Concretely, evaluation metrics for vulnerable examples are $56.4\%$/$69.6\%$ EM/F$_1$, compared to $73.0\% / 76.5\%$ on the whole dataset (-16.6\% and -6.9\% EM/F$_1$).

Attackable questions have on average $12.3$ tokens, whereas unattackable ones are slightly shorter with on average $11.1$ tokens. 
We considered the distribution of different question types (\emph{What, Who, When, ...}) for both attackable and unattackable samples and did not observe notable differences apart from the single most frequent question type \emph{What}; it is a lot more prevalent among the unattacked questions (56.4\%) than among successfully attacked questions (42.1\%). 
This is by far the most common question type, and furthermore one that is comparatively open-ended and does not prescribe particular type expectations to its answer, as e.g., a \emph{Where} question would require a location.
A possible explanation for the prevalence of the \emph{What} questions among unsuccessfully attacked samples is thus, that the model cannot rely on type constraints alone to arrive at its predictions \cite{sugawara2018what}, and is thus less prone to such exploitation. Section \ref{sec:biased} will study this in more detail.

Finally, Figure~\ref{fig:ner_histogram} shows a histogram of the $10$ most common NE tags appearing in unsuccessfully attacked samples versus the corresponding fraction of replaced entities in successfully attacked samples.
Besides one exception, the distributions are remarkably similar. Undersensitivity can be induced for a variety of entity types used in the perturbation, but in particular questions with geopolitical entities (\emph{GPE}) are error-prone. 
A possible explanation can be provided by observations regarding (non-contextualised) word embeddings, which cluster geopolitical entities (e.g.~countries) close to one another, thus making them potentially hard to distinguish for a model operating on these embeddings \citep{Mikolov2013distributed}.
%

\begin{table*}[t]
\begin{center}
\begin{tabular}{@{\extracolsep{2pt}}l rrrr   rr r  rr@{}}
\toprule
 {\bf{SQuAD2.0}}&\multicolumn{4}{c}{ \bf{Undersensitivity Error Rate}}&\multicolumn{2}{c}{\bf{HasAns}}&\multicolumn{1}{c}{\bf{NoAns}}&\multicolumn{2}{c}{\bf{Overall}}\\
 \cline{2-5}  \cline{6-7}   \cline{8-8}  \cline{9-10}
Adv.~budget $\eta$ & ~$@32$  & ~$@64$ & $@128$ & $@256$  & {EM} & {F$_1$}   & EM/F1& EM  & F$_1$ \\
\midrule
\textsc{BERT Large}     & 44.0      &  50.3     & 52.7      & 54.7          & \bf{70.1} & \bf{77.1}     & 76.0 &  73.0     &   76.5        \\
\quad + Data Augment.   & {\bf 4.5} &  {\bf9.1} & {\bf 11.9}&  {\bf 18.9}   & 66.1 & 72.2     & {\bf80.7}&  {\bf73.4}&   76.5        \\
\quad + Adv. Training   &  11.0     & 15.9      &  22.8     & 28.3          & 69.0 & 76.4     & 77.1 &  73.0     &  {\bf 76.7}   \\
\bottomrule
\end{tabular}
\end{center}
\caption{Breakdown of undersensitivity error rate overall (lower is better), and standard performance metrics (EM, F$_1$; higher is better) on different subsets of \emph{SQuAD2.0} evaluation data, all in [\%].} \label{tab:defense_results}
\end{table*}

\begin{table*}[t]
\begin{center}
\begin{tabular}{@{\extracolsep{2pt}}l rrrr   rr r  rr@{}}
\toprule
 {\bf{NewsQA}} &\multicolumn{4}{c}{ \bf{Undersensitivity Error Rate}}&\multicolumn{2}{c}{\bf{HasAns}}&\multicolumn{1}{c}{\bf{NoAns}}&\multicolumn{2}{c}{\bf{Overall}}\\
 \cline{2-5}  \cline{6-7}   \cline{8-8}  \cline{9-10}
Adv.~budget $\eta$ & ~$@32$  & ~$@64$ & $@128$ & $@256$  & {EM} & {F$_1$}   & EM/F1& EM  & F$_1$ \\
\midrule
\textsc{BERT Base}      & 34.2   & 34.7  & 36.4  &  37.3    & {\bf41.6}  & 53.1   &  61.6 & 45.7  &  54.8   \\
\quad + Data Augment.   & {\bf7.1} & {\bf11.6} & {\bf17.5}  & {\bf20.8}  & 41.5 & {\bf53.6}  & 62.1 & {\bf45.8} &  {\bf55.3}       \\
\quad + Adv. Training   & 20.1 & 24.1 & 26.9 & 29.1 & 39.0 &50.4  &  {\bf67.1} &  44.8 & 53.9 \\
\bottomrule
\end{tabular}
\end{center}
\caption{Breakdown of undersensitivity error rate overall (lower is better), and standard performance metrics (EM, F$_1$; higher is better) on different subsets of \emph{NewsQA} evaluation data, all in [\%].} \label{tab:defense_results_newsqa}
\end{table*}

\section{Defending Against Undersensitivity Attacks}
We will now investigate methods for mitigating excessive model undersensitivity.
Prior work has considered both data augmentation and adversarial training for more robust models, and we will conduct experiments with both.
Adding a robustness objective can negatively impact standard test metrics \citep{tsipras2018robustness}, and it should be noted that there exists a natural trade-off between performance on one particular test set and performance on a dataset of adversarial inputs.
We perform data augmentation and adversarial training by adding a corresponding loss term to the standard log-likelihood training objective:
\begin{equation}
    \mathcal{L}^{Total} = \mathcal{L}^{llh}(\Omega) + \lambda \cdot \mathcal{L}^{llh}(\Omega')
\end{equation}
where $\Omega$ is the standard training data, fit with a discriminative log-likelihood objective,  $\Omega'$ either a set of augmentation data points, or of successful adversarial attacks where they exist, and $\lambda>0$ a hyperparameter.
In data augmentation, we randomly sample perturbed input questions, whereas in adversarial training we perform an adversarial search to identify them.
In both cases, alternative data points in $\Omega'$ will be fit to a \emph{NULL} label to represent the \emph{NoAnswer} prediction -- again using a log-likelihood objective.
Note that we continuously update $\Omega'$ throughout training to reflect adversarial samples w.r.t. the current model.

\paragraph{Experimental Setup: \emph{SQuAD2.0}}
We train the \textsc{BERT Large} model on \emph{SQuAD2.0}, tuning the hyperparameter $\lambda \in \{ 0.0, 0.01, 0.1, 0.25,$
$0.5, 0.75, 1.0, 2.0\}$, and find $\lambda=0.25$ to work best for both of the two defence strategies.
We tune the threshold for predicting \emph{NoAnswer} based on validation data and report results on the test set (the original \emph{SQuAD2.0} Dev set). 
All experiments are executed with batch size 16, NE perturbations are used for the defence methods, and adversarial attacks with $\eta=32$ and $\rho=1$ in adversarial training.
Where no attack is found for a given question sampled during SGD training, we instead consider a different sample from the original training data.
We evaluate the model on its validation data every 5,000 steps and perform early stopping with a patience of 5.

\paragraph{Experimental Setup: \emph{NewsQA}}
Following the experimental protocol for \emph{SQuAD}, we further test a \textsc{BERT Base} model on \emph{NewsQA} \citep{trischler2017newsqa}, which -- like \emph{SQuAD2.0} -- contains unanswerable questions.
As annotators do often not fully agree on their annotation in \emph{NewsQA}, we opt for a conservative choice and filter the dataset, such that only samples with the same majority annotation are retained, following the preprocessing pipeline of \citet{talmor2019multiqa}.

\paragraph{Experimental Outcomes}
Results for these experiments can be found in Table~\ref{tab:defense_results} and Table~\ref{tab:defense_results_newsqa} for the two datasets, respectively.
First, we observe that both data augmentation and adversarial training substantially reduce the number of undersensitivity errors the model commits, consistently across adversarial search budgets, and consistently across the two datasets.
This demonstrates that both training methods are effective defences and can mitigate -- but not eliminate -- the model's undersensitivity problem.
Notably the improved robustness -- especially for data augmentation -- is possible without sacrificing performance in the overall standard metrics EM and F$_1$, even slight improvements are possible.

Second, data augmentation is a more effective defence training strategy than adversarial training.
This holds true both in terms of standard and adversarial metrics, and hints potentially at some adversarial overfitting on the training set.
Finally, a closer inspection of how performance changes on answerable (\emph{HasAns}) vs.~unanswerable (\emph{NoAns}) samples of the datasets reveals that models with modified training objectives show improved performance on unanswerable samples, while sacrificing some performance on answerable samples.\footnote{Note that the \emph{NoAns} prediction threshold is fine-tuned on the respective validation sets.}
This suggests that the trained models -- even though similar in standard metrics -- evolve on different paths during training, and the modified objectives prioritise fitting unanswerable questions to a higher degree.

\begin{table*}[t]\label{tab:biased_lewisfan}
\begin{center}
\begin{tabular}{@{\extracolsep{3pt}} l cc cc cc @{}}
\toprule
\multicolumn{1}{c}{}&\multicolumn{2}{c}{{\bf Person}} &\multicolumn{2}{c}{ {\bf Date} }&\multicolumn{2}{c}{{\bf Numerical}}\\
 \cline{2-3}  \cline{4-5}  \cline{6-7}
            & {EM} & {F$_1$}  & {EM} & {F$_1$}  & {EM} & {F$_1$}  \\
\midrule
\textsc{BERT Base} - w/ data bias   & 55.9    & 63.1      & 48.9     &  58.2    &  38.7    & 48.0     \\
\quad + Robust Training  &{\bf 59.1}& {\bf 66.6} & {\bf 58.4} & {\bf 65.6} & {\bf 48.7} & {\bf 58.9} \\
\midrule
\textsc{BERT Base} - w/o data bias  &  69.2   &  78.1      &   73.2    &   81.7&   69.6   & 80.5    \\
\bottomrule
\end{tabular}
\end{center}
\caption{Robust training leads to improved generalisation under train/test distribution mismatch (data bias, top). Bottom: control experiment without train/test mismatch. } \label{tab:results_biased_data}
\end{table*}

\subsection{Evaluation on Held-Out Perturbation Spaces}\label{section:new_perturbations}
In Tables \ref{tab:defense_results} and \ref{tab:defense_results_newsqa} results are computed using the same perturbation spaces also used during training. 
These perturbation spaces are relatively large, and questions are about a disjoint set of articles at evaluation time. Nevertheless there is the potential of overfitting to the particular perturbations used during training.
To measure the extent to which the defences generalise also to new, held out sets of perturbations, we assemble a new, disjoint perturbation space of identical size per NE tag as those used during training, and evaluate models  on attacks with respect to these perturbations.
Named entities are chosen from English Wikipedia using the same method as for the training perturbation spaces, and chosen such that they are disjoint from the training perturbation space.
We then execute adversarial attacks using these new attack spaces on the previously trained models, and find that both vulnerability rates of the standard model, as well as relative defence success transfer to the new attack spaces.
For example, with $\eta=256$ we observe vulnerability ratios of 51.7\%, 20.7\%, and 23.8\% on \emph{SQuAD2.0} for standard training, data augmentation, and adversarial training, respectively. Detailed results for different values of $\eta$, as well as for \emph{NewsQA} can be found in Appendix \ref{apdx:perturbation_generalisation}.

\subsection{Generalisation in a Biased Data Setting}\label{sec:biased}
Datasets for high-level NLP tasks often come with annotation and selection biases; models then learn to exploit shortcut triggers which are dataset- but not task-specific \citep{jia2017adversarial,DBLP:conf/naacl/GururanganSLSBS18}.
For example, a model might be confronted with question/paragraph pairs which only ever contain one type-consistent answer span, e.g.~mention one number in a text with~a \emph{How many...?} question.
It is then sufficient to learn to pick out numbers from text to solve the task, irrespective of other information given in the question. 
Such a model might then have trouble generalising to articles that mention several numbers, as it never learned that it is necessary to take into account other relevant question information that helps determine the correct answer.

We test models in such a scenario:
a model is trained on \emph{SQuAD1.1} questions with paragraphs containing only a single type-consistent answer expression for either a person, date, or numerical answer.
At test time, we present it with question/article pairs of the same respective question types, but now there are \emph{multiple} possible type-consistent answers in the paragraph.
We obtain such data from \citet{lewis2018generative}, who first described this biased data scenario.
Previous experiments on this dataset were conducted without dedicated development set, so while using the same training data, we split the test set with a 40/60\% split\footnote{Approximate, as we stratify by article.} into development and test data.\footnote{We also include an experiment with the previous data setup used by \citet{lewis2018generative}, see Appendix \ref{apdx:lewis_fan}. }
We then test both a vanilla fine-tuned \textsc{BERT Base} transformer model, and a model trained to be less vulnerable to undersensitivity attacks using data augmentation.
Finally, we perform a control experiment, where we join and shuffle all data points from train/dev/test (of each question type, respectively), and split the dataset into new parts of the same size as before, which now follow the same data distribution (w/o data bias setting).

Table~\ref{tab:results_biased_data} shows the results.
In this biased data scenario we observe a marked improvement across metrics and answer type categories when a model is trained with unanswerable samples (robust training).
This demonstrates that the negative training signal stemming from related -- but unanswerable -- questions counterbalances the signal from answerable questions in such a way, that the model learns to better take into account relevant information in the question, which allows it to correctly distinguish among several type-consistent answer possibilities in the text, which the standard \textsc{BERT Base} model does not learn well.

\subsection{Evaluation on Adversarial SQuAD}
We next evaluated \textsc{BERT Large} and \textsc{BERT Large} + Augmentation Training on \textsc{AddSent} and \textsc{AddOneSent}, which contain adversarially composed samples \citep{jia2017adversarial}.
Our results, summarised in Table~\ref{tab:transfer} in the Appendix, show that including altered samples during the training of \textsc{BERT Large} improves EM/F$_1$ scores by 2.7/3.7 and 0.1/1.6 points on the two datasets, respectively.

\subsection{Transferability of Attacks}
We train a RoBERTa model \citep{liu2019roberta} on \emph{SQuAD2.0}, and conduct undersensitivity attacks ($\rho=6$, $\eta=256$). 
For the same attack budget, error rates are considerably lower for RoBERTa (34.5\%) than for BERT (54.7\%). 
When considering only samples where RoBERTa was found vulnerable, BERT also has a vulnerability rate of 90.7\%. 
Concrete adversarial inputs $\vecx'$ chosen for RoBERTa transfer when evaluating BERT for 17.5\% of samples (i.e.~satisfy Inequality \ref{eq:undersensitivity}).

\section{Conclusion}
We have investigated a problematic behaviour of RC models -- being overly stable in their predictions when given semantically altered questions.
We find that model robustness to undersensitivity attacks can be drastically improved using appropriate defences, such as data augmentation and adversarial training, without sacrificing in-distribution test set performance.
Future work should address in more detail the causes and better defences to model undersensitivity, which we believe can provide an alternative viewpoint on a model's RC capabilities.

\subsubsection*{Acknowledgments}
This research was supported by an Engineering and Physical Sciences Research Council (EPSRC) scholarship.

\bibliography{reference_list}
\bibliographystyle{acl_natbib.bst}
\clearpage

\appendix

\begin{table}[t]
\footnotesize
\begin{center}
\begin{tabular}{@{\extracolsep{2pt}}l rrrr  @{}}
\toprule
 {\bf{SQuAD2.0}}&\multicolumn{4}{c}{ \bf{Undersensitivity Error Rate}}\\
 \cline{2-5}  
Adv.~budget $\eta$ & ~$@32$  & ~$@64$ & $@128$ & $@256$   \\
\midrule
\textsc{BERT Large}         & 40.7 & 45.2   & 48.6  & 51.7      \\
\quad + Data Augment.       & {\bf 4.8}  & {\bf 7.9}    & {\bf 11.9}  & {\bf 20.7}      \\
\quad + Adv. Training       & 9.2  & 12.2   & 16.5  & 23.8      \\
\bottomrule
\end{tabular}
\end{center}
\caption{Breakdown of undersensitivity error rate on \emph{SQuAD2.0} with a held-out attack space (lower is better).} \label{tab:defense_transfer_results_squad}
\end{table}

\begin{table}[t]
\footnotesize
\begin{center}
\begin{tabular}{@{\extracolsep{2pt}}l rrrr  @{}}
\toprule
 {\bf{NewsQA}}&\multicolumn{4}{c}{ \bf{Undersensitivity Error Rate}}\\
 \cline{2-5}  
Adv.~budget $\eta$ & ~$@32$  & ~$@64$ & $@128$ & $@256$   \\
\midrule
\textsc{BERT Base}          & 32.8  & 33.9  & 35.0  & 36.2      \\
\quad + Data Augment.       & \bf{3.9}   &  \bf{6.5}  & \bf{11.9}  & \bf{17.5}      \\
\quad + Adv. Training       & 17.6  & 20.7  & 25.4  & 28.5      \\
\bottomrule
\end{tabular}
\end{center}
\caption{Breakdown of undersensitivity error rate on \emph{NewsQA} with a held-out attack space (lower is better).} \label{tab:defense_transfer_results_newsqa}
\end{table}

\begin{figure}[t]
    \centering
    \includegraphics[width=1.0\columnwidth]{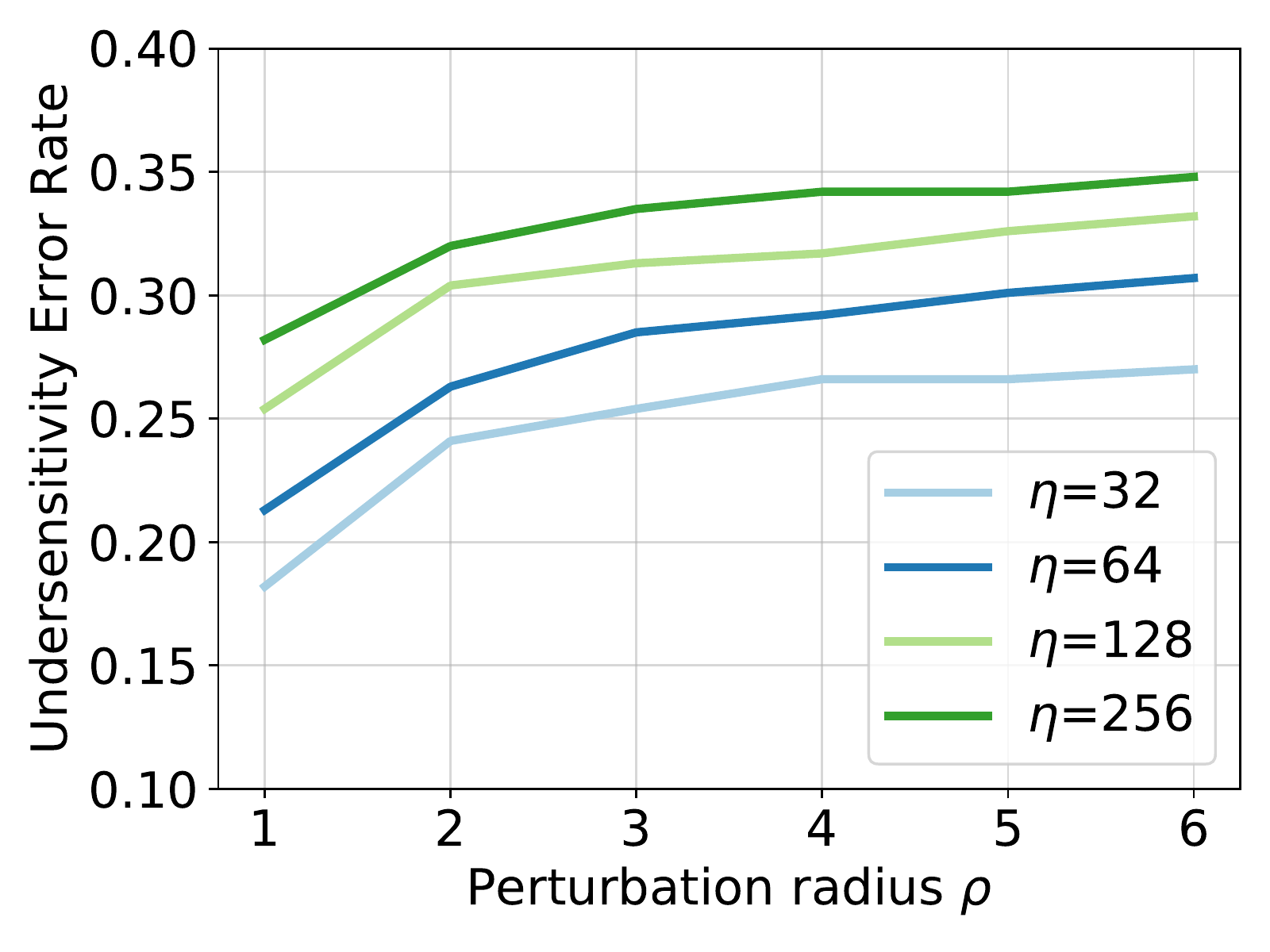}
    \caption{Vulnerability to undersensitivity attacks on \emph{NewsQA}.} 
    \label{fig:vulnerability_newsqa}
\end{figure}

\section{Appendix: PoS Perturbation Details.}\label{apdx:pos}
We exclude these PoS-tags when computing perturbations: \emph{`IN', `DT', `.', `VBD', `VBZ', `WP', `WRB', `WDT', `CC', `MD', `TO'}.

\section{Appendix: Generalisation to Held-out Perturbations}\label{apdx:perturbation_generalisation}
Vulnerability results for new, held-out perturbation spaces, disjoint from those used during training, can be found in Table \ref{tab:defense_transfer_results_squad} for \emph{SQuAD2.0}, and in Table~\ref{tab:defense_transfer_results_newsqa} for \emph{NewsQA}.

\section{Appendix: Adversarial Example from a Question Collection}\label{apdx:example}
Searching in a large collection of (mostly unrelated) natural language questions, e.g.~among all questions in the \emph{SQuAD2.0} training set, yields several cases where the prediction of the model increases, compared to the original question, see Table~\ref{tab:alternative_question} for one example. 
Such cases are however rare, and we found the yield of this type of search to be very low.

\section{Appendix: Attack Examples}
Table~\ref{tab:labeled_attacks_appendix} shows more examples of successful adversarial attacks on \emph{SQuAD2.0}.

\begin{table*}[t]
\begin{center}
\begin{tabular}{ll}
\toprule
{\bf Given Text}    & [...] The Normans were famed for their martial spirit and eventually for their \\
                    & Christian piety, becoming exponents of the {\emph{Catholic orthodoxy}} [...] \\
{\bf Q (orig)}      &   What religion were the Normans? \stbox{0.78} \\
{\bf Q (adv.)}      & IP and AM are most commonly defined by what type of proof system? \itbox{0.84}\\
\bottomrule
\end{tabular}
\end{center}
\caption{Drastic example for lack of specificity: unrelated questions can trigger the same prediction (here: \emph{Catholic orthodoxy}), and  with higher probability.}\label{tab:alternative_question}
\end{table*}

\begin{table*}[t]
\centering
\begin{tabular}{l rr c rr c rr }
\toprule
\multicolumn{1}{c}{} & \multicolumn{2}{c}{\textsc{AddSent}} & & \multicolumn{2}{c}{\textsc{AddOneSent}} & & \multicolumn{2}{c}{\textsc{Dev 2.0}} \\
\cline{2-3} \cline{5-6} \cline{8-9}
& \textbf{EM} & \textbf{F$_1$} & & \textbf{EM} & \textbf{F$_1$} & & \textbf{EM} & \textbf{F$_1$} \\
\midrule
BERT Large & 61.3 & 66.0 & & 70.1 & 74.9 & & 78.3 & 81.4 \\
BERT Large+NE defence & \textbf{64.0} & \textbf{70.3} & & \textbf{70.2} & \textbf{76.5} & & \textbf{78.9} & \textbf{82.1} \\
\bottomrule
\end{tabular}
\caption{Comparison between \textsc{BERT Large} and \textsc{BERT Large} + data augmentation using NE perturbations, on two sets of adversarial examples: \textsc{AddSent} and \textsc{AddOneSent} from \citet{jia2017adversarial}.} \label{tab:transfer}
\end{table*}

\begin{table*}[t]
\begin{center}
\begin{tabular}{l l l r r}
\toprule
{\bf Original / Modified Question}  & \bf{Prediction} & {\bf{Annotation}} & \bf{Scores} \\
\midrule
What ethnic neighborhood in {\stbox{Fresno}} {\itbox{Kilbride}} had  & Chinatown & valid  & \stbox{0.998}   \\
primarily Japanese residents in 1940? & & &                                                \itbox{0.999}    \\
\midrule
The {\stbox{Mitchell Tower}} \itbox{MIT} is designed to look & Magdalen & valid & \stbox{0.96} \\ 
 like what Oxford tower?   & Tower & & \itbox{0.97}  \\
\midrule
\midrule
What does the EU's {\stbox{legitimacy}} {\itbox{digimon}} rest on?  & the ultimate & valid & \stbox{0.38} \\
& authority of [...] & &     \itbox{0.40} \\
\midrule
What is Jacksonville's hottest recorded  &104$^{\circ}$F & valid & \stbox{0.60} \\
{\stbox{temperature}} {\itbox{atm}}?  &&&\itbox{0.62} \\
\bottomrule
\end{tabular}
\end{center}
\caption{Example adversarial questions (\stbox{original}, \itbox{attack}), together with their annotation as either a valid counterexample or other type.  Top 2: Named entity (NE) perturbations. Bottom 2: PoS perturbations.}\label{tab:labeled_attacks_appendix}
\end{table*}

\begin{table*}[t]\label{tab:biased_lewisfan2}
\begin{center}
\begin{tabular}{@{\extracolsep{3pt}} l C{1cm}C{1cm} C{1cm}C{1cm} C{1cm}C{1cm}@{}}
\toprule
\multicolumn{1}{c}{}&\multicolumn{2}{c}{{\bf Person}} &\multicolumn{2}{c}{ {\bf Date} }&\multicolumn{2}{c}{{\bf Numerical}}\\
 \cline{2-3}  \cline{4-5}  \cline{6-7}
            & {EM} & {F$_1$}  & {EM} & {F$_1$}  & {EM} & {F$_1$}  \\
\midrule
GQA \citep{lewis2018generative}     &  53.1     &  61.9     &   64.7    &   72.5    & {\bf 58.5}& {\bf 67.6}     \\
\midrule
\textsc{BERT Base} - w/ data bias   & 66.0      & 72.5      & 67.1      &  72.0     &  46.6     & 54.5     \\
\quad + Robust Training             & {\bf 67.4}&{\bf 72.8} & {\bf 68.1}&{\bf 74.4} & 56.3      & 64.5 \\
\bottomrule
\end{tabular}
\end{center}
\caption{Robust training leads to improved generalisation under train/test distribution mismatch (data bias).} \label{tab:lewis_fan_original}
\end{table*}

\section{Appendix: Biased Data Setup}\label{apdx:lewis_fan}
For completeness and direct comparability, we also include an experiment with the data setup of \citet{lewis2018generative} (not holding aside a dedicated validation set).
Results can be found in Table~\ref{tab:lewis_fan_original}. 
We again observe improvements in the biased data setting. 
The robust model outperforms GQA~\citep{lewis2018generative} in two of the three subtasks.
%
\section{Appendix: Vulnerability Analysis on NewsQA}
Figure~\ref{fig:vulnerability_newsqa} depicts the vulnerability of a \textsc{BERT Large} model on \emph{NewsQA} under attacks using NE perturbations.

\end{document}